\pdfoutput=1
\documentclass{article}
\usepackage{graphicx}
\usepackage{times}
\usepackage{url}
\usepackage{amsmath,amssymb} 
\usepackage{color}
\usepackage[scriptsize]{subfigure}
\usepackage{bbm}
\usepackage{multirow}
\usepackage{authblk}

\usepackage[width=122mm,left=12mm,paperwidth=146mm,height=193mm,top=12mm,paperheight=217mm]{geometry}
\usepackage{hyperref}

\begin{document}
\newcommand*{\email}[1]{%
	\normalsize\href{mailto:#1}{#1}\par
}

\title{Real-time 3D Deep Multi-Camera Tracking} 

\author{Quanzeng You}
\author{Hao Jiang} 
\affil{Microsoft Cloud \& AI, One Microsoft Way, Redmond, WA \\ \email{quanzeng.you, jiang.hao@microsoft.com}}
\date{}
\maketitle

\begin{abstract}
Tracking a crowd in 3D using multiple RGB cameras is a challenging task.
Most previous multi-camera tracking algorithms are designed for offline setting and 
have high computational complexity.
Robust real-time multi-camera 3D tracking is still an unsolved problem. In this work, 
we propose a novel end-to-end tracking pipeline, Deep Multi-Camera Tracking (DMCT), which achieves
reliable real-time multi-camera people tracking. 
Our DMCT consists of 1) a fast and novel perspective-aware Deep GroudPoint Network,
2) a fusion procedure for ground-plane occupancy heatmap estimation, 
3) a novel Deep Glimpse Network for person detection and
4) a fast and accurate online tracker.  Our design fully unleashes the power of deep neural network to
estimate the ``ground point'' of each person in each color image, which can be optimized to run efficiently and robustly.  Our fusion procedure,
glimpse network and tracker merge
the results from different views, find people candidates using multiple video frames and then track people on the fused heatmap. 
Our system achieves the state-of-the-art tracking results while maintaining real-time performance.
Apart from evaluation on the challenging WILDTRACK dataset, we also collect two more tracking datasets with high-quality labels from two different environments and camera settings. 
Our experimental results confirm that our proposed real-time pipeline gives superior results to previous approaches.
\end{abstract}

\section{Introduction}
Tracking a crowd in a large space is a challenging task, which
has been receiving a lot of interests since it is the basis for people behavior understanding and has
numerous applications in retail, surveillance and manufacture industry.
Single camera tracking is hard to cover an
extended space and has difficulty in dealing with crowd and occlusion.
Multiple overlapping cameras can be used to cover an almost unlimited space.
When designed properly, multi-camera tracking can handle clutter and occlusion and
achieve much higher accuracy than single camera tracking.

However, multi-camera tracking has its own unique challenges. It requires us to maintain
the consistency of each target's identity across multiple views.
When there is a large number of cameras, the efficiency of the tracking algorithm becomes critical.
Most traditional multi-camera tracking algorithms have been designed for offline tracking and have high complexity.
Therefore, they are hard to be deployed in real-time applications.
Tracking a crowd reliably through a large space in clutter using multiple cameras 
is still unsolved.
In this paper, we propose a novel real-time pipeline, Deep Multi-Camera Tracking (DMCT),
to tackle this problem.

\begin{figure}[!tbp]
	\begin{center}
		\includegraphics[width=0.95\linewidth]{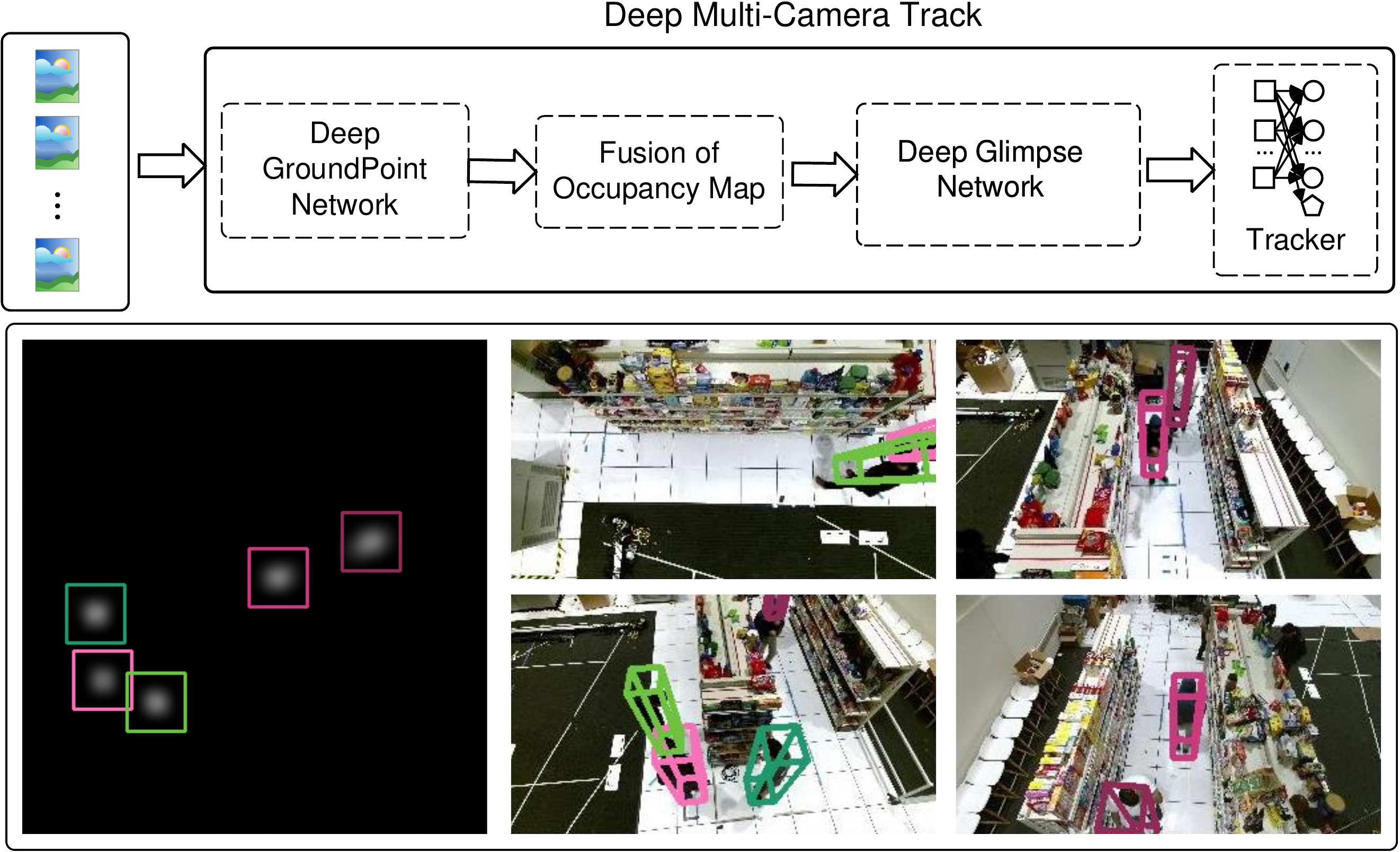}
	\end{center}
	\caption{End-to-end real-time 3D deep multi-camera tracking (DMCT). \textbf{First row}:
   Multiple live video streams pass our proposed \textit{perspective aware Deep GroundPoint Network},
	 \textit{fusion procedure}, \textit{Deep Glimpse Network for people detection}, and \textit{fast online tracker}
	 to track multiple people in 3D in real time.
	  \textbf{Second row} shows example tracking result on the fused occupancy map (column 1) and the projected 3D bounding boxes on each RGB view (columns 2 and 3) in a cluttered environment.}
	\label{fig:pipeline}
\end{figure}

\figurename~\ref{fig:pipeline} illustrates the framework of the proposed pipeline.
We design a Deep GroundPoint Network to estimate the ``ground point'', 
which is the projection of the weight center of each person onto the virtual ground plane.
In particular, our novel designs take the perspective projection effect into consideration when estimating the ground point heatmap in each camera view. 
The deep ground-point network is applied to images from each camera to generate ground point probability maps.     
Next, the probability maps from all the views are fused on a shared ground plane using 3D geometry, which is
also known as the occupancy map~\cite{epfl1}. We build a novel 
light-weighted people detection module, Deep Glimpse Network, on the fused occupancy map.
The Deep Glimpse Network  takes a sequence of occupancy maps and employs glim\textit{}pse layer and 
temporal convolutional layers for real-time and robust person detection. 

In the end, our fast online tracker links the person candidate detections into trajectories. \figurename~\ref{fig:pipeline} illustrates
a tracking result of our proposed method, where we show the top-down view of each tracked person on the ground plane. We also show
the projected \textbf{3D bounding boxes} with a fixed width and height for each tracked person in all camera views. 
Our contributions include:
\begin{itemize}
	\item We develop a novel perspective-aware learning objective for detecting people's ``ground points'' in each camera view.
	\item We propose a simple, yet effective candidate proposal method on the fused occupancy map, which is constructed by fusing the projected groundpoint heatmaps from all camera views.  
	\item We propose a novel Deep Glimpse Network, which attempts to capture the motions for better people recognition results.
	\item Our proposed method achieves end-to-end real-time online tracking, while at the same time outperforms
	previous state-of-the-art offline tracking methods.
	\item We also collect two multi-camera people tracking datasets with two different settings. High quality ground-truth labels are provided and the datasets will be released to the research community.
\end{itemize}

\section{Related works}
Multi-camera multi-target tracking has been intensively studied.
Existing approaches can be generally classified into two categories.
Early works, such as \cite{singletrack1}, track targets in
each single view and a correspondence algorithm is used to match targets from different views to maintain the identity consistency.
In~\cite{medioni}, 
a top-down view
occupancy map is used for jointly tracking the 2D bonding boxes in each camera view and the 3D locations on the top-down map.
However, single camera
tracking often loses track due to occlusions, especially
when dealing with crowd and clutter.
Frequent occlusions cause fragmented tracklets in each camera view.
Fusion of these fragmented tracklets from multi-camera views to generate consistent 3D tracking result is
a non-trivial task. Different optimization methods \cite{opt1,opt2,opt3} have been proposed to tackle
this problem. These methods take advantage of the geometrical constraints
among sensors and use the deep features to re-identify people across multiple
views. These methods often have high complexity and may need multiple passes to process the data;
it is hard to achieve real-time online tracking using these approaches.

The other strategy of multi-camera tracking uses a centralized
representation instead of dealing with each separate view.
The occupancy
map (2D) and occupancy volume (3D) method belong to this approach. Our proposed method follows such a theme.
In dense camera settings, space carving \cite{count}
can be used to generate a 3D occupancy volume, where target objects can be extracted using background
subtraction or semantic segmentation. This method has been used in \cite{count} for people tracking and counting.
However, it
may generate ``ghost'' objects especially when there is a small number of cameras and there is a crowd.

In addition, the occlusion relationships among different subjects have also been explicitly modeled
for more robust people tracking \cite{occ}.
In \cite{centroid}, only the centroid of each target foreground map from background subtraction
is used when constructing the
occupancy maps. This method requires at least two cameras that can see each target so that triangulation can be used to
find the 3D location.
Another triangulation method for tracking a large number of bats has also been proposed in \cite{margrit}.
 Still, ghost targets may appear during triangulation.
In \cite{larry}, foreground pixels in each
camera view are matched to reconstruct their 3D positions. The 3D point cloud is then projected to the ground plane
to generate occupancy map
for people
tracking. Ground plane homograph \cite{ucf} has also be used to generate a voting map from the foreground pixels in each view
for occupancy map construction. Probabilistic method \cite{epfl1} has been proposed to generate more robust occupancy maps
and sequential dynamic programming method is used in people tracking.

More recently, a deep learning method~\cite{epfl2} is proposed for people detection and occupancy map construction. 
This method has high
complexity and it is also unclear whether it can be used on top-down view images. In \cite{count19}, a seven-layer
fully convolution network is proposed for constructing people probability maps and people counting.

Comparing with tracking targets in each single view, the occupancy map approach does not need to handle the identity consistency problem
across different views. However, constructing a high-quality occupancy map and then tracking people efficiently and effectively using such a
representation
is still
an open problem. In this paper, we propose a deep learning approach to solve the problem.
Our DMCT network is designed to work on multiple frames simultaneously in videos.
Our task of estimating people's ``ground point'' is similar to keypoints detection in computer vision~\cite{alahi2012freak,leutenegger2011brisk}, which has been extensively studied.
Recently, CornerNet \cite{law2018cornernet} and its variants CenterNet \cite{duan2019centernet,zhou2019objects} achieve competing performance on object detection task.
One essential component in these frameworks is the estimation of keypoints for candidate objects.
Inspired by these works, we construct our perspective aware ``ground point'' estimation network.
In this way, we can build accurate occupancy maps for our tracker.

Indeed, our method can be applied to
any videos from arbitrary cameras, either top-down or side-view or a mixture, as confirmed by our experiments, where we can even train from ``pseudo ground point'' from large
datasets such as CrowdHuman~\cite{shao2018crowdhuman} and deploy the model in our lab.

\section{Method}
We use multiple calibrated RGB cameras for multi-people tracking. These cameras may have small overlaps. In fact, our
method still works if there is only one camera.  
We fully employ the powerful deep neural network to analyze and understand the visual content in each single camera view;
on the other hand,
we also make use of 3D geometry, which is known to be difficult for current deep neural networks to model, to merge
the results from different cameras.

\begin{figure*}[!htbp]
	\begin{center}
		\includegraphics[width=0.95\linewidth]{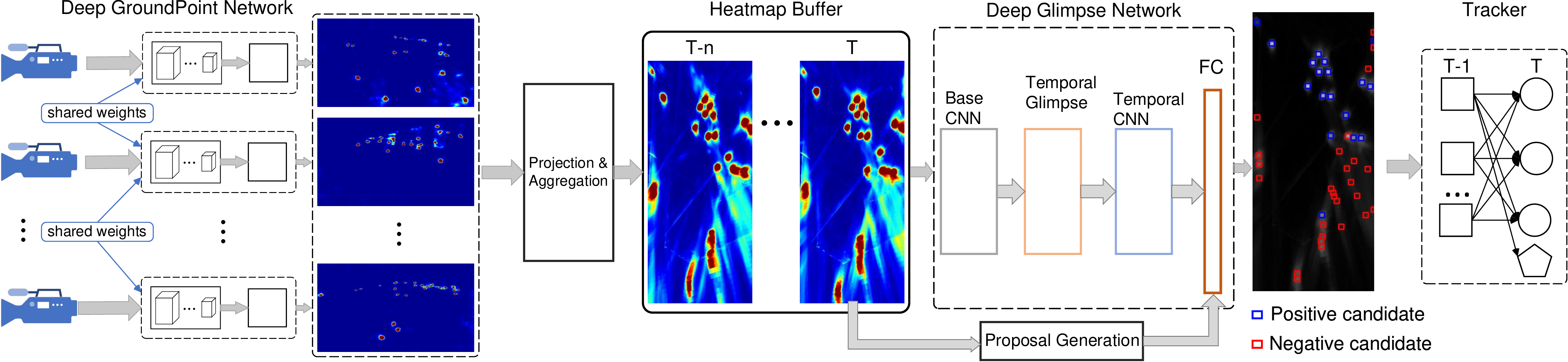}
	\end{center}
	\caption{
		Our 3D Deep Multi-Camera Tracking pipeline.
	}
	\label{fig:dmct}
\end{figure*}

\figurename~\ref{fig:dmct} illustrates the framework of the proposed pipeline.
Our DMCT reads RGB images from multiple calibrated cameras. A perspective-aware Deep GroundPoint Network is
employed to estimate the ``ground point''. To avoid the projection distortion~\cite{count19}, we design a perspective-aware loss to generate
more accurate heatmap estimations on the shared top-down view.

Next, we use the camera matrices to project the view-wise heatmaps to construct
the ground-plane occupancy map. The projected heatmaps from multiple cameras can then be averaged or stacked as input to our Deep Glimpse Network.
Our algorithm, which is constructed to process the ground heatmaps and features from color images of all the cameras, 
will benefit from less occlusions, measurable distance space, invariance to camera view angles and less computational resources required.
The detection results at each time instant are the inputs to our tracker, which mainly uses the estimated person ground locations to associate
targets through time. We also take advantage of the view-wise RGB images by projecting each detected person to
 each view (see \figurename{~\ref{fig:pipeline}} for an example). More expensive approaches, such as DNN, can be used to extract visual features for each candidate. However, as shown in our experiments,
 even simple RGB histograms are favorable for our tracker. The details of different modules are discussed in the following sections.

\subsection{Perspective aware groundpoint estimation}
\label{sec:heatmap}


\begin{figure}
	\begin{center}
		\subfigure[Input image]{
			\includegraphics[width=0.225\linewidth]{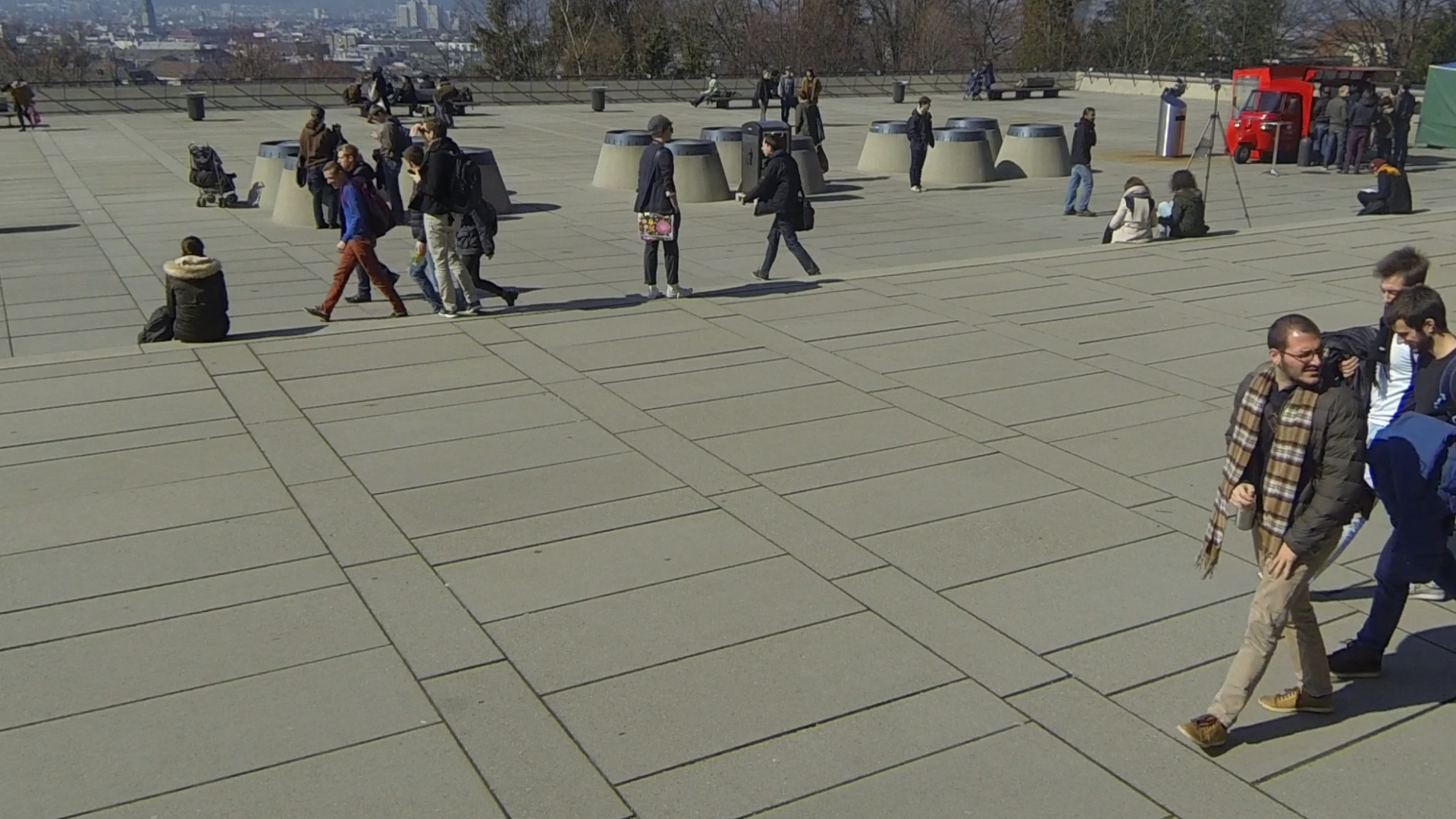}
			\label{fig:obj:ipt}
		}
		\subfigure[Mask map]{
			\includegraphics[width=0.225\linewidth]{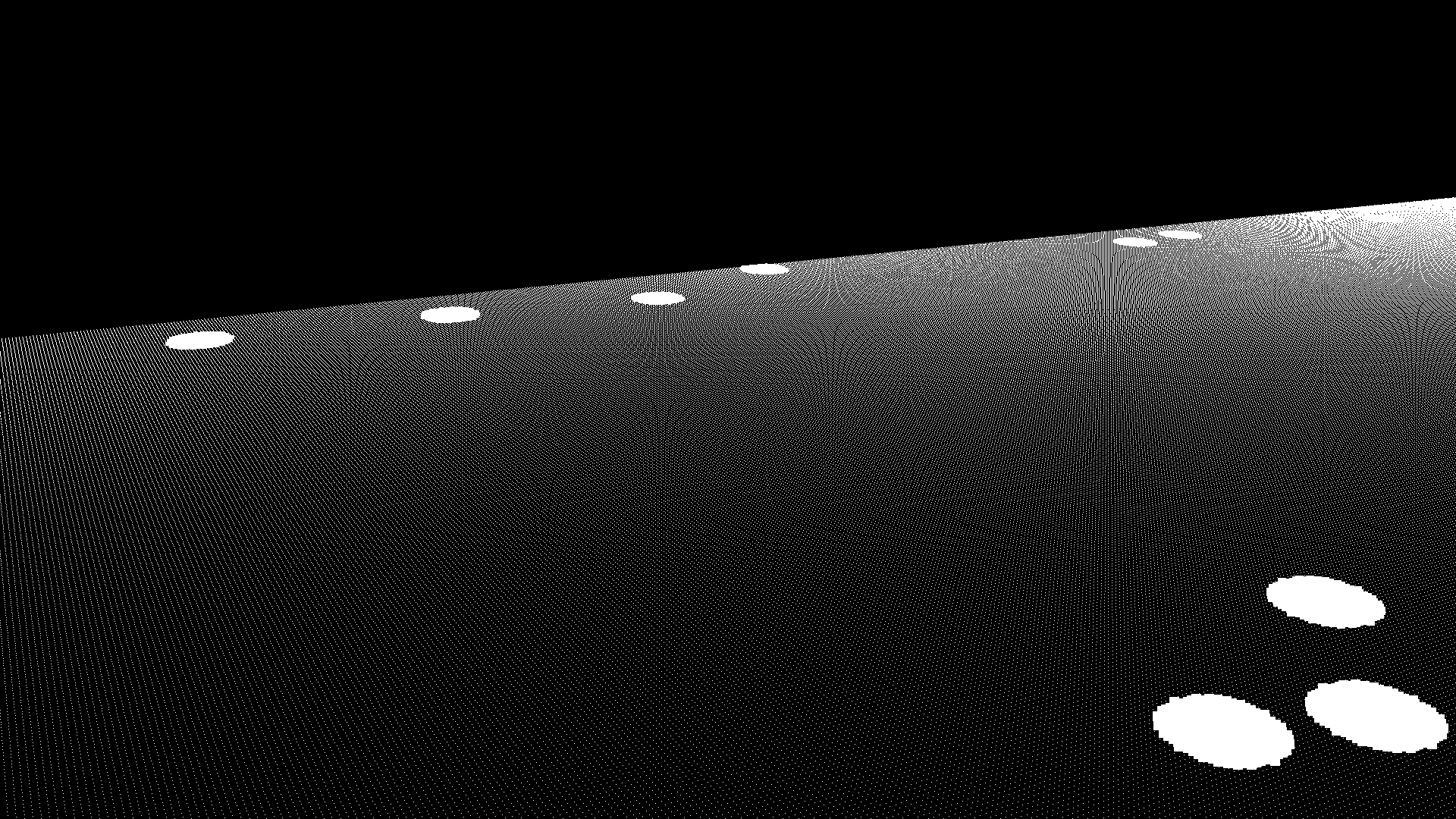}
			\label{fig:obj:mask}
		}
		\subfigure[Ground-truth map]{
			\includegraphics[width=0.225\linewidth]{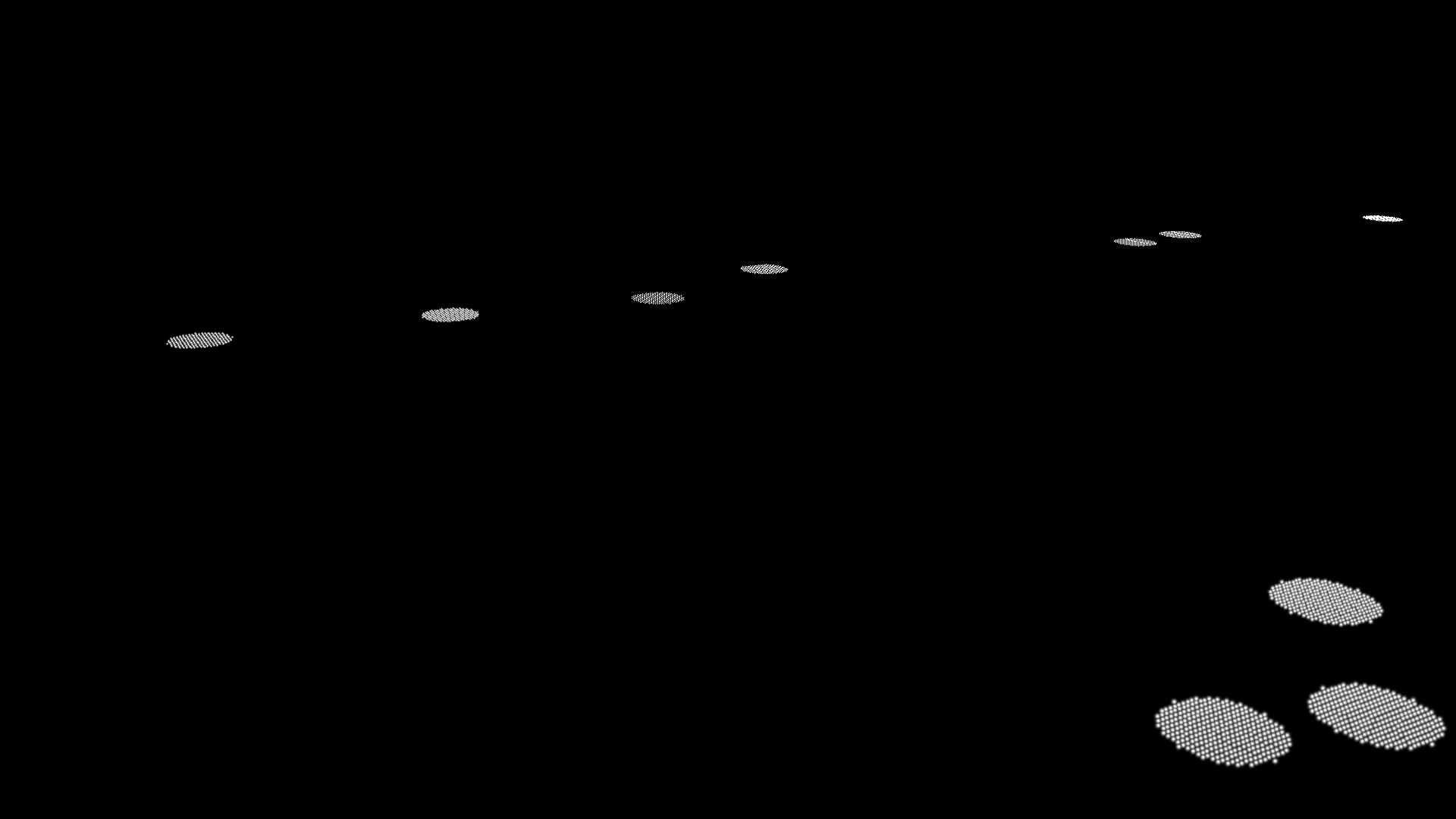}
			\label{fig:obj:hm}
		}
		\subfigure[Gaussian heatmap]{
			\includegraphics[width=0.225\linewidth]{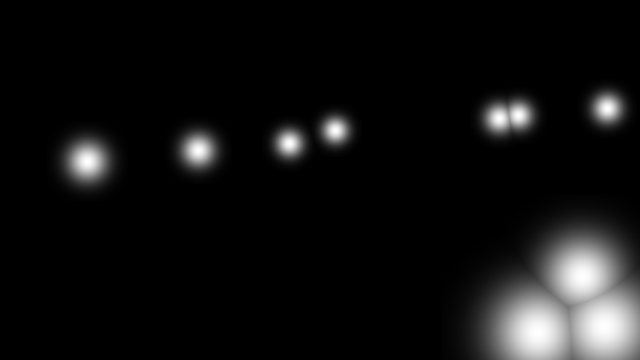}
			\label{fig:obj:gaussian}
		}
	\end{center}
	\caption{Construction of the learning objective for groundpoints detection. (d) is widely used by CornerNet and its variants as ground-truth heatmap.
	Our method using (b) and (c) corrects the perspective effect, while (d) cannot. }
	\label{fig:obj}
\end{figure}

\textbf{Deep GroundPoint network} CornerNet and its variants employ a Gaussian kernel to construct ground-truth heatmap $H$. For each ground-truth ``ground point'' $(i_t, j_t)$, a scale-aware radius
$r_t$ is estimated. For each location $(i,j)$ within the circle at $(i_t, j_t)$ with a radius of $r_t$, 
we set $H_{ij} = \exp\left(-\frac{(i - i_t)^2 + (j - j_t)^2}{2\sigma_t^2}\right)$, where we follow~\cite{law2018cornernet} and set $\sigma_t$ to be one third of  $r_t$. If location $(i,j)$ is within the coverage of multiple ``ground points'', we set $H_{ij}$ to the maximum of all the computed values.  
\figurename{~\ref{fig:obj:gaussian}} shows an example of the ground-truth heatmap constructed in this approach.
This learning target shows competing performance on 2D object detection. However, it is inappropriate for our task.

In our task, we need to map person location heatmap from camera views to the virtual
ground plane. Gaussian heatmap (see \figurename{~\ref{fig:obj:gaussian}}) used in previous method does not take into consideration the
perspective transformation and thus
leads to undesired elongated distortion during the mapping, which complicates
the following people detection. To solve this problem,
when generating ground truth occupancy map on each image, we enforce that each person on the ground plane occupies a disk with equal radius and we
back project the person occupancy map to each color image to form the view-wise ground truth heatmap.
We therefore ``pre-distort'' the occupancy map on each image to correct the perspective distortion
on the ground-plane occupancy map.
\figurename{~\ref{fig:obj:hm}} shows an example of the projected map. 
We discretize the ground plane into cells of $2.5 cm$ and project all the grid points to form the mask. 
We use the mask in the loss function.
\figurename{~\ref{fig:obj:mask}} illustrates the mask map for the given example input image in \figurename{~\ref{fig:obj:ipt}}.

We apply the pixel-wise Focal Loss~\cite{law2018cornernet} to train our groundpoint
detection model. Given the  binary mask $M$ and the heatmap $H$ whose elements are in $[0,1]$, our perspective aware learning objective is
\begin{equation}
L = -\frac{1}{N} \sum_{ij} \left\{
\begin{split} & M_{ij}(1-P_{ij})^\alpha \log(P_{ij}) \;\; \mathit{if} H_{ij} = 1\\
&\begin{split}M_{ij}(1-H_{ij})^\beta \\ P_{ij} ^\alpha \log(1-P_{ij})\end{split}  \qquad \quad \;\mathit{otherwise}
\end{split}
\right.
\label{eqn:loss}
\end{equation}
where $P$ is the predicted heatmap, $N$ is  the number of non-zero elements in mask $M$, $\alpha$ and $\beta$ are hyper-parameters in Focal Loss.

\subsubsection{Heatmap fusion onto the ground plane}
To fuse the predicted heatmaps $\{P_1, P_2, \cdots, P_C\}$ ($C$ is the number of cameras), we compute the homographies $\{R_1, R_2, \cdots, R_C\}$ between the shared ground plane and all camera views. We use two methods for heatmap fusion. 
In method one, we average projected heatmaps to produce the final occupancy map $\bar{P}$:

\begin{equation}
I_{ij}  = \sum_{c=1}^C \mathbbm{1}( R^c_{ij} ), \;
\bar{P}_{ij} = \frac{\sum_{c=1}^C \mathbbm{1}( R^c_{ij} ) P^c_{R^c_{ij}} } {I_{ij}},
\end{equation}
where $R^c_{ij}$ computes the projected ground-plane coordinate for the $c$-th view location $(i,j)$  and $\mathbbm{1}(\cdot)$ returns one if $R^c_{ij}$ is valid coordinate on
$c$-th view otherwise it returns zero.
In method two, we stack all heatmaps to generate inputs for our person detector. Method two makes the person detector depend on the number of cameras.
In our experiments, we compare the performance of our tracker using both methods.

\subsection{People tracking on fused heatmaps}
\label{sec:tracking}
Our tracking method first detects target candidates and then links them into continuous trajectories.
Inspired by the success of R-CNN~\cite{Girshick_2014_CVPR} and its fast successor Faster R-CNN~\cite{ren2015faster} on object detection,
we build a two-stage people detector on the fused heatmaps. Its results are the inputs to our online tracker.
\subsubsection{Light-weighted candidate proposal generation}
\begin{figure}
	\begin{center}
	\subfigure[Ground-truth locations]{
		\includegraphics[width=0.45\linewidth]{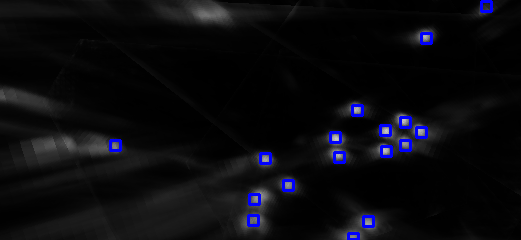}
		\label{fig:heatmap:gt}
	}
	\subfigure[Generated proposals]{
	\includegraphics[width=0.45\linewidth]{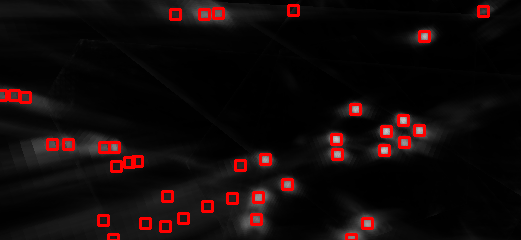}
	\label{fig:heatmap:det}
	}
	\end{center}
   	\caption{Detection proposal from local maxima: (a) blue bounding boxes represent the ground-truth people locations and (b) red bounding boxes show the proposal candidates corresponding to the local maxima.}
	\label{fig:heatmap}
\end{figure}

Ideally, the extracted occupancy heatmap would be binary and perfectly match
the ground-truth people occupancy on the ground plane, where we have probability of one for the locations where people occupy and probability of zero for all other locations. However, due to errors introduced by the prediction model and the noises
from the projections, each element on these heatmaps has value in $[0,1]$. One example is shown in \figurename~{\ref{fig:heatmap:gt}}, where each blue bounding box shows the ground-truth location of each person.

After careful examination of \figurename{~\ref{fig:heatmap:gt}}, we observe that each ground-truth bounding box corresponds to a blob in the probability map.
Therefore, instead of forming a computationally expensive deep neural network, such as Region Proposal Network~\cite{ren2015faster}, to generate the proposal candidates for detection, we generate proposals by finding all the local maxima on the heatmap (see \figurename{~\ref{fig:heatmap:det}} for an example of all local maxima).

\begin{figure*}
	\begin{center}
		\subfigure[Architecture of our Deep Glimpse Network]{
			\includegraphics[width=0.525\linewidth]{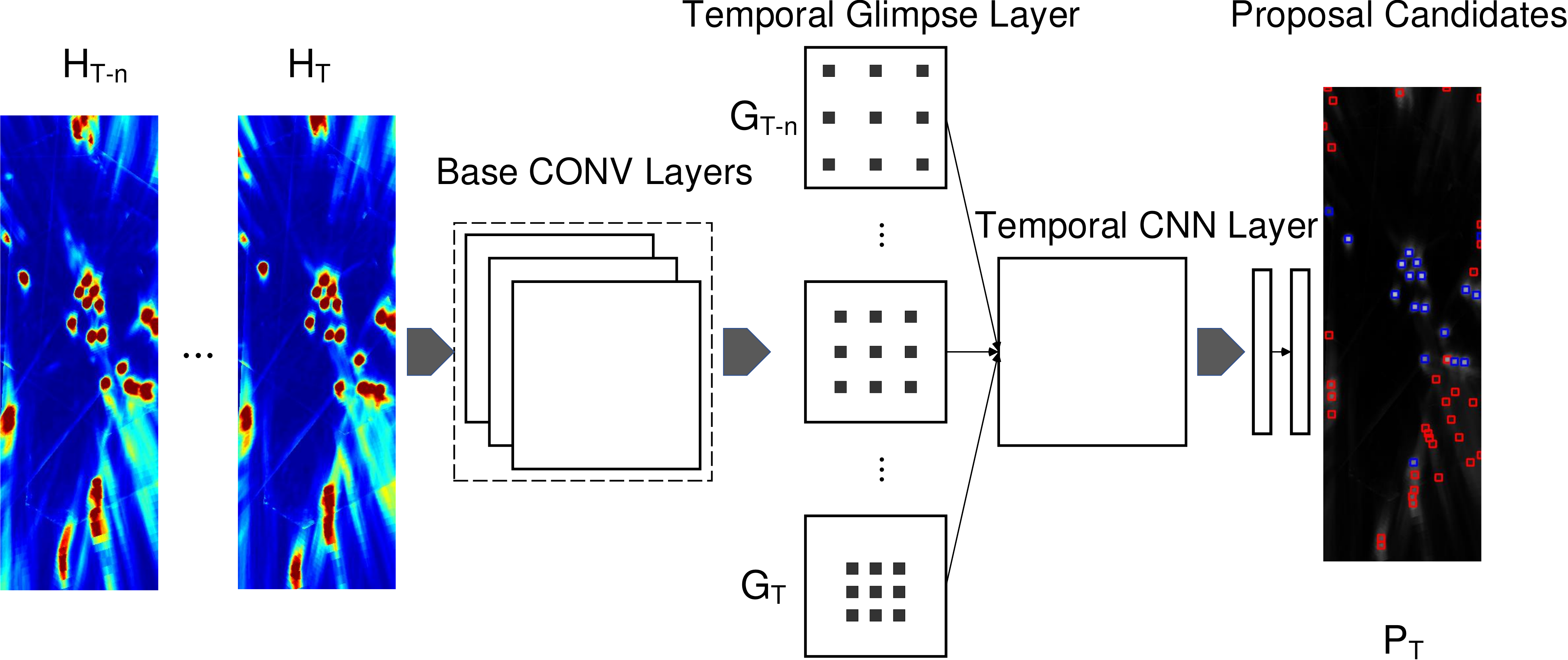}
			\label{fig:model:person}
		}%
		\hspace{1cm}%
		\subfigure[Online Tracker]{
			\includegraphics[width=0.215\linewidth]{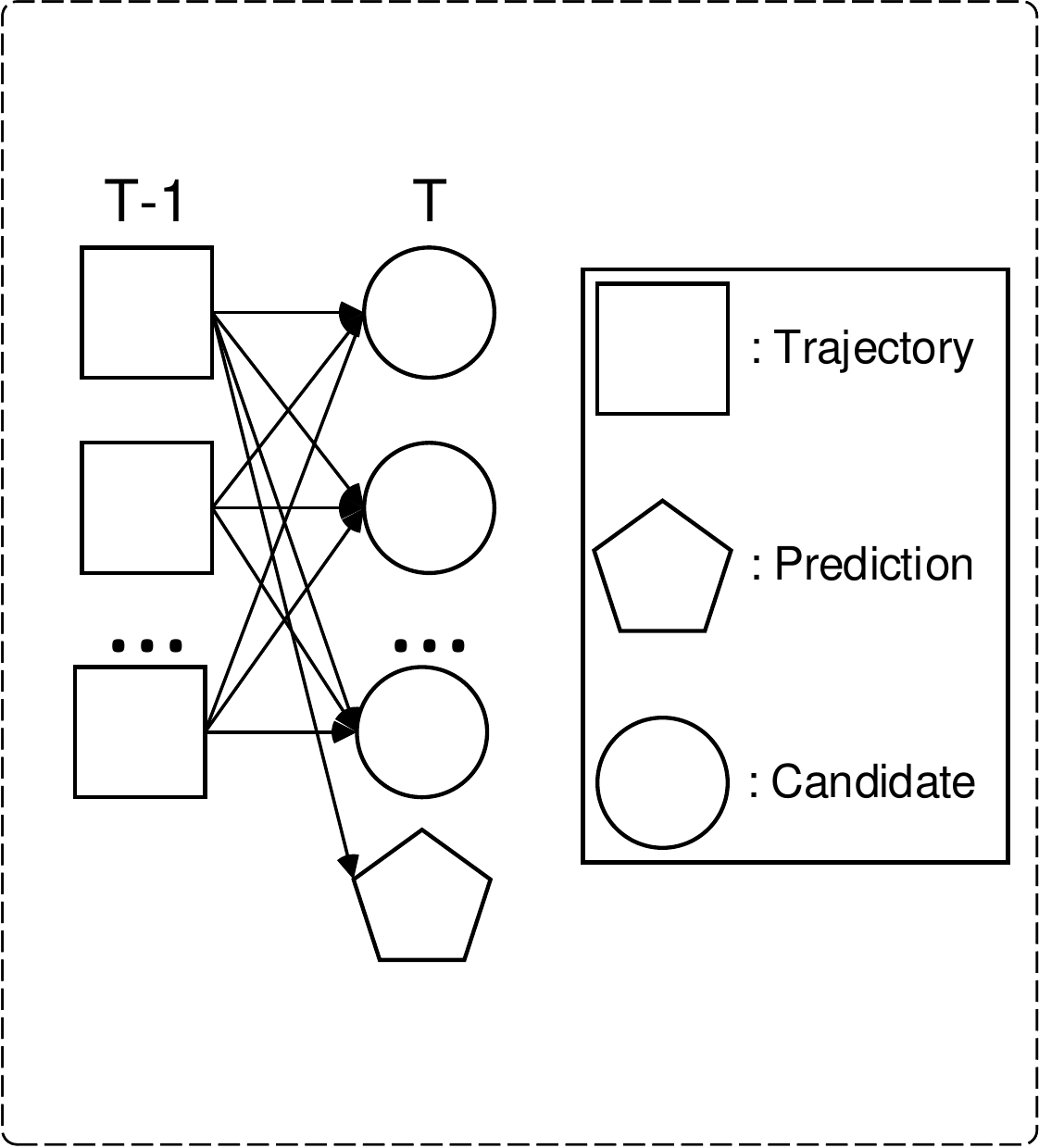}
			\label{fig:peoplenet}
		}%
	\end{center}
	\caption{Architecture of our Deep Glimpse Network and our online tracker. The network consists of several base convolutional layers, a temporal glimpse layer, a temporal CNN layer and two fully-connected layers (see main text for details). }
	
\end{figure*}
\subsection{Deep Glimpse Network for person recognition}
As shown in \figurename{~\ref{fig:heatmap}}, our candidate proposals have high recall but
give many false alarms as well.
We thus further process each proposal candidate to classify them into people and non-people categories. 
Following similar strategy in \cite{Girshick_2014_CVPR},
we compute the IoU $I_x$ between each proposal $x$ and its nearest ground-truth candidate.
We use the IoU to label each candidate proposal and build a deep neural network for the recognition task. More specifically, we label proposal $x$ as ``positive'' class if $I_x$ is above a threshold $I_T$. Otherwise, we label it as ``negative'' class.

The ``noisy'' proposals can, mostly, be attributed to the relatively large projection errors
introduced by the large distance between these regions and the cameras. We expect the neural network model can learn from the heatmap patterns to recognize positive and
negative detections.
We employ a temporal sequence to reduce the impact of ``noisy'' regions.
Intuitively, we anticipate temporal sequence could provide people's movements as additional information to facilitate neural network training.
Inspired by the design of \textit{spatial} glimpse sensor in~\cite{mnih2014recurrent}, we propose a \textit{temporal} glimpse sensor to learn from sequence of probability maps.
To some extent, the temporal glimpse sensors simulate the ``zoom in'' process when human eyes capture the movement in a video. Thus, the model is offered with
more temporal context and is expected to be robust against noisy candidate proposals.

The architecture of our recognition model is shown in \figurename~\ref{fig:model:person}. We use current heatmap $H_T$ to generate the candidate proposals.
Then, we take the previous $n$ heatmaps, along with the current heatmap, to build the people recognition model. All heatmaps share several base convolutional
layers to construct low-level feature maps. Next, the sequence of feature maps is forwarded through different convolutional layers.
Earlier feature maps use layers with larger dilations. We call them ``temporal glimpse layer''.
A temporal convolutional layer~\cite{bai2018empirical} is employed to fuse the outputs from the previous glimpse layer to produce the final pixel-wise feature maps.
We employ the features at the center location of each candidate proposal to build the person classifier. Compared with other 2D detection models,
our model assumes a fixed size bounding box because people have the same scales from top-down view. This further simplifies the design of our neural networks,
where extra operations, such as ROI pooling, ROI Align and NMS, have been safely removed.
\subsection{Real-time people tracking}
With the extracted candidates, people tracking can be formulated as a path following problem.
We try to link the detected trajectories at time $T-1$ to the detections in the current frame $T$.
The tracking graph is shown in \figurename{~\ref{fig:peoplenet}}: the rectangle nodes represent
the trajectories already formed, the oval nodes represent candidates detected, and the pentagon nodes are the prediction nodes. The number of prediction nodes
equals the number of rectangular nodes at previous time instant. 
The edges indicate possible matches between nodes.
The edge weights are determined by the Euclidean distance\footnote{As shown in our experiments, color information can also be introduced and benefit our tracker.}.
To track objects in the scene, we find the extension of each trajectory from time $T-1$ to
$T$, so that these paths pass each trajectory node and all the paths are node disjoint.

The tracking problem, where we associate the paths in the above constraints, can be reduced to a
min-cost max-flow problem
and
can be solved efficiently using a polynomial algorithm \cite{papadimitrou1982combinatorial}.
Each trajectory is only extended to the
neighboring nodes within a radius $d_L$, which is determined by the max speed
of a person and the frame rate of the tracking algorithm.
After the optimization, we extend each existing trajectory by one-unit length. We remove
trajectories which have no matched candidates in a pre-defined number of steps, \textit{e.g.} 100. 
 And, we include new trajectory for each
candidate node at time $T$ that is not on any path. The new set of trajectories are used to form
a new graph for the next time instant.
\section{Experiments}
We conduct two separate experiments to validate the effectiveness of the proposed pipelines: 1) on the challenging WILDTRACK~\cite{chavdarova2018wildtrack}  dataset and 2) our new datasets from two different environments.
WILDTRACK dataset is a challenging dataset that captures videos in a real-world setting. Even though our approach is an online tracker, it demonstrates better performance than the
 offline trackers on WILDTRACK as shown in the following experimental results. Our new datasets are also challenging; they involve more varieties of different camera angles and strong clutter.
\subsection{Results on WILDTRACK dataset}
We follow the supervised learning settings in~\cite{chavdarova2018wildtrack}, where we use the last $10\%$ of the labeled data as testing data 
and the first $90\%$ of the labeled data are used for training and validating.

\begin{figure}[!htbp]
	\begin{center}
		\includegraphics[width=0.85\linewidth]{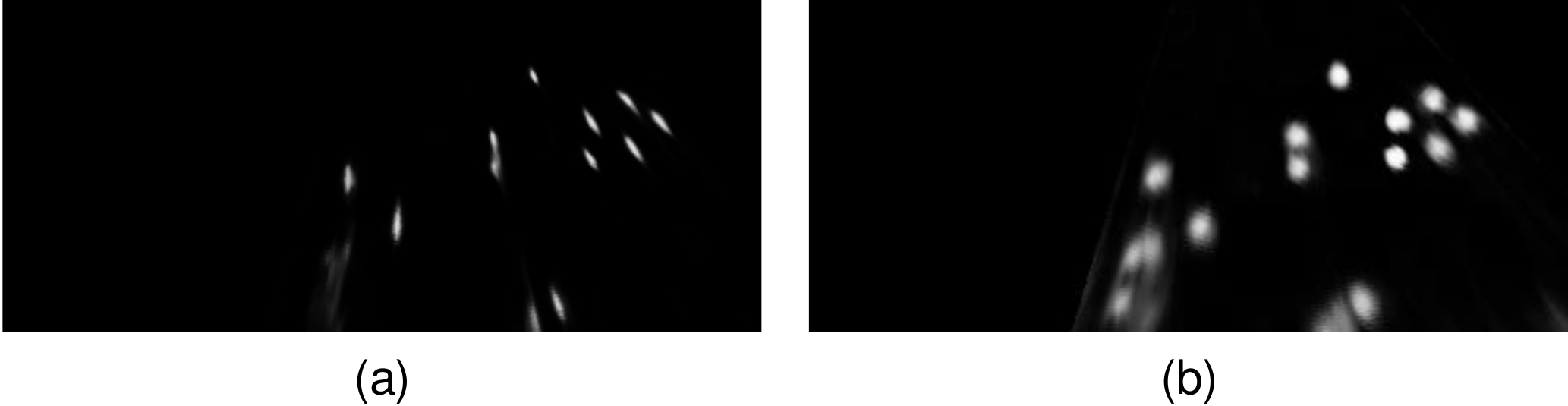}
	\end{center}
	\caption{Examples of the projected heatmap on the ground plane: (a) the results using
	the method in \cite{zhou2019objects} and (b) our results. }
	\label{fig:feet:pred}
\end{figure}

\textbf{Ground point prediction on each view}
We use Eq.~(\ref{eqn:loss}) as the loss function to train our Deep GroundPoint Network for ``ground point'' estimation on each image. 
We use the Deep Layer Aggregation (DLA)~\cite{yu2018deep} network as
the backbone, which
has a good balance between efficiency and accuracy.
In our implementation, we use an output stride of two to better predict groundpoints for far and small persons in each view.
\figurename{~\ref{fig:feet:pred}} shows some examples of the projected heatmaps on the ground plane. The first heatmap shows the
results from the model trained following \cite{zhou2019objects} which also uses the DLA network
as the
backbone. And the second heatmap is the prediction results of our model.
Our method maps each candidate to a desired disk-like shape with similar radius
as demonstrated by the examples in~\figurename{~\ref{fig:feet:pred}}.
Due to the limit of image resolution, as the distance from a camera to a target becomes
very large, the occupancy map becomes increasingly blurry. A simple peak detection method is not sufficient
for robust people detection.
We therefore need our proposed glimpse network
to achieve accurate results.

\begin{table}[!htbp]
	\begin{center}
		\caption{Performance of different person classification models on the testing data: accuracy, precision, recall and weighted F1 score.}
		\label{tab:people:det}
		\begin{tabular}{l|l|l|l|l}
			\hline
			Model &	Accuracy	&	Precision	&	Recall	&	F1	\\  \hline
			T-CNN	&	93.07\%	&	93.08\%	&	93.07\%	&	93.06\%	\\
			T-CNN Stack	&	93.22\%	&	93.22\%	&	93.22\%	&	93.22\%	\\
			T-Glimpse	&	93.52\%	&	93.58\%	&	93.52\%	&	93.49\%	\\
			T-Glimpse Stack	&	\textbf{94.79\%}	&	\textbf{94.82\%}	&	\textbf{94.79\%}	&	\textbf{94.80\%}	\\
			\hline
		\end{tabular}
	\end{center}
\end{table}

\textbf{Performance of person detection}
We compare the performance of different variants of the person recognition model on the occupancy probability map: 1) \textbf{T-CNN} is our temporal
convolutional neural network, it uses convolutional layers with the same dilation; 2) \textbf{T-CNN Stack} accepts the predicted heatmaps $\{P_1, P_2, \cdots, P_C\}$ as a multi-channel input instead of using the
averaged heatmap $\bar{P}$ as in \textbf{T-CNN}; 3) \textbf{T-Glimpse} uses the temporal glimpse layer to learn the pixel-wise features and 4)~\textbf{T-Glimpse Stack} is the same with \textbf{T-CNN Stack}, which also
uses view-wise heatmaps as input.

The results for different variants of our model are shown in \tablename{~\ref{tab:people:det}}. Using view-wise heatmaps as inputs demonstrates superior performance to its counterparts using averaged
heatmaps. This can be potentially due to the neural network's ability of automatically learning the weights for different views when we fuse them into a unified heatmap.
In addition, the temporal glimpse layer exhibits better results than temporal CNN under both stacked heatmaps and averaged heatmaps settings.

\textbf{Results of our online tracker}
Our online tracker works on the fused occupancy heatmap and accepts our person detection results as inputs.
We primarily use the locations to compute the associating weights. We also experiment with tracker using the color histogram to
demonstrate the effect of colors to our system. To compute the color histogram, we assume a
two-meter high cuboid with a width of $60\mathit{cm}$. We center the cuboid at each detected person
ground location and then project the cuboid to each view, then fit a 2D bounding box to the projected cuboid (see~\figurename{~\ref{fig:pipeline}} for examples of projected cuboid).
We accumulate each person's color histogram within the fitted bounding box in all the views.

\begin{table*}

	\begin{center}
		\caption{Multiple people tracking results on the WILDTRACK dataset.}
		\label{tab:tracking:wl}
			\scriptsize{
		\begin{tabular}{l|p{0.3cm}p{0.3cm}p{0.3cm}p{0.3cm}p{0.3cm}p{0.3cm}p{0.3cm}p{0.3cm}p{0.3cm}p{0.5cm}p{0.5cm}}
			\hline
			Method	&	IDF1	&	IDP	&	IDR	&	MT	&	ML	&	FP	&	FN	&	IDS	&	FM	&	MOTA	&	MOTP	\\ \hline
			DeepOcclusion+KSP	&	73.2	&	83.8	&	65	&	28.7\%	&	25.1\%	&	1095	&	7503	&	85	&	92	&	69.6	&	61.5	\\
			DeepOcclusion+KSP+ptrack	&	78.4	&	84.4	&	73.1	&	42.1\%	&	14.6\%	&	2007	&	5830	&	103	&	95	&	72.2	&	60.3	\\
			T-Glimpse (ours)	&	77.8	&	79.3	&	76.4	&	61\%	&	\textbf{4.9\%}	&	\textbf{91}	&	126	&	42	&	43	&	72.8	&	\textbf{79.1}	\\
			T-Glimpse Stack (ours)	&	\textbf{81.9}	&	\textbf{81.6}	&	\textbf{82.2}	&	\textbf{65.9\%}	&	\textbf{4.9\%}	&	114	&	\textbf{107}	&	\textbf{21}	&	\textbf{34}	&	\textbf{74.6}	&	78.9	\\
			\hline
		\end{tabular}
		}
	\end{center}
\end{table*}
The tracking results are shown in \tablename~\ref{tab:tracking:wl}. We use the open-sourced \textbf{py-motmetrics} (\url{https://github.com/cheind/py-motmetrics}) library to compute all the metrics
for benchmarking multiple object trackers. The same benchmarking tool (Matlab version) and benchmarking method are also used in \cite{chavdarova2018wildtrack}.

The results are computed with a default IoU threshold of $0.5$ and we use a squared bounding box with the edge length of one meter (the same setting as~\cite{chavdarova2018wildtrack}) for each candidate on the ground plane.
The results show that our online tracker achieves better performance. 
Especially, our tracker
shows better results in terms of MOTA, MOTP and IDF1, which are mostly interested to the MOT task.
Note that our tracker is real-time and online, while the competing methods here are offline
methods. Our method still gives better results despite the disadvantage.

\begin{table}
	\begin{center}
		\caption{Comparisons between different variants of our method. }
		\label{tab:tracking:wlablation}
		\begin{tabular}{l|p{0.4cm}p{0.4cm}p{0.7cm}p{0.8cm}|p{0.4cm}p{0.4cm}p{0.7cm}p{0.8cm}}
			\hline
			&\multicolumn{4}{c}{With color histogram} & \multicolumn{4}{|c}{Without color histogram} \\ \hline
Method	&	IDF1	&	IDs	&	MOTA	&	MOTP&	IDF1	&	IDs	&	MOTA	&	MOTP	\\ \hline
T-CNN	&	70.3	&	39	&	70.7	&	79&	68.2	&	45	&	70.2	&	79.1	\\
T-CNN Stack	&	68.4	&	34	&	73.1	&	78.9&	65.5	&	47	&	71.4	&	78.9	\\
T-Glimpse	&	77.8	&	42	&	72.8	&	\textbf{79.1}&	\textbf{70.6}	&	50	&	\textbf{72.1}	&	\textbf{79.1}	\\
T-Glimpse Stack	&	\textbf{81.9}	&	\textbf{21}	&	\textbf{74.6}	&	78.9&	68.8	&	\textbf{43}	&	71.6	&	79	\\ \hline
		\end{tabular}
	\end{center}
	
\end{table}

\textbf{Ablation study} Our ablation study results are summarized in~\tablename~{\ref{tab:tracking:wlablation}}. 
We use the same parameters for all the different variants of our method. 
The differences come from
the person detection model (mainly the person recognition model as shown in~\tablename{~\ref{tab:people:det}}). The temporal glimpse layer greatly improves the result. This is particularly true for the score of IDF1 and MOTA.  As shown in~\tablename~{\ref{tab:tracking:wlablation}}, 
the extra color information improves the identification F-score (IDF1) and reduces ID switches.
A more powerful deep neural network to represent color information could give even better results.

\subsection{Results on our collected datasets}
\begin{figure}[!htbp]
	\begin{center}
	\subfigure[Env1]{
		\includegraphics[width=0.4\linewidth]{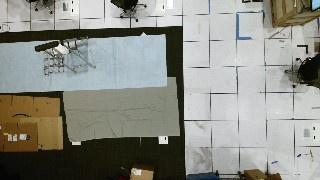}
	}
	\subfigure[Env2]{
		\includegraphics[width=0.4\linewidth]{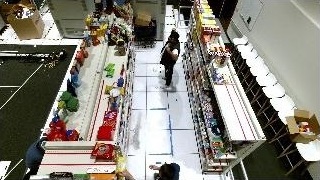}
	}
	\end{center}
	\caption{Example images from the two environments. We only show one camera view from each environment.}
	\label{fig:studiox}
\end{figure}

To further investigate the effectiveness of the proposed solutions, we also collect our own datasets in two different settings and evaluate our models with several baselines.
In the first setting, we install 4 overhead Kinect V2 cameras with some furniture in the space. In
the second setting, we deploy 4 Kinect V2 cameras to a different space, which has two rows of shelves with a variety of different products.
\figurename{~\ref{fig:studiox}} shows two example images from the two environments with one of the 4 cameras.

In Env1, we collect several different sequences of videos: training and testing involve different subjects. 
In Env2, we use the same group of subjects, but with changed clothing in training and testing sequences.
All video sequences are recorded with an approximate 15 frames per second. The depth images are used to generate the ground truth for each frame and the ground truth tracking results for all
sequences.
Only RGB images are used for training and testing. All cameras are calibrated and we use the best-effort synchronization to collect data from all connected cameras. 

We have 12 subjects and 5 subjects for the data collection in Env1 and Env2 respectively. 
In total, we collect $18,414 \times 4$ images for training and validating in Env1 and $2,525 \times 4$ images for testing. 
In Env2, we have a small dataset with $2,001 \times 4$ for training and validating, and $1,301 \times 4$ images for testing. 
We compare our approach with the following baselines:
\begin{itemize}
\item {\textbf{VGG}: We use VGG-16~\cite{simonyan2014very} as our backbone network to learn a pixel-wise segmentation classifier on each color image. 
The fully convolutional network has a structure similar to the edge detection network in ~\cite{HED}.
The learning objective is to segment the ``ground points'' in each camera view.}
	\item{\textbf{ResNet}: We apply the same setting as the above \textbf{VGG}, but using ResNet-50~\cite{he2016deep} as the backbone.}
	\item{\textbf{FCN7}}: We follow~\cite{count19} to employ the FCN7 backbone to train a view-wise heatmap and then use the same setting as~\textbf{VGG} to construct the tracker.
	\item{\textbf{YOLO}: Pre-trained YOLO~\cite{yolo} detector has poor performance on our environments. This is particularly true for the overhead cameras (Env 1).  In this work,
		we therefore train a YOLO ``ground point'' detector for each view using the ground truth.
                The detector finds a bounding box centered at each person ground point.  
}
\end{itemize}

Since all baselines are trained without a person recognition module, we report all results without applying the Deep Glimpse Network for further filtering.
The view-wise heatmaps for our approach along with all the baselines are fused into a ground-plane heatmap using the same procedure.
Next, we build our tracker on the fused ground-plane heatmaps using the same hyper-parameters for all approaches. All these settings offer fair comparisons between all
methods.
\begin{table}[!tbp]
	\begin{center}
		{
		\caption{Comparisons between different approaches for Env 1. Overall, ours show the best performance. }
		\label{tab:tracking:env1}
		\begin{tabular}{l|p{0.4cm}p{0.4cm}p{0.4cm}p{0.2cm}p{0.6cm}p{0.8cm}|p{0.4cm}p{0.4cm}p{0.4cm}p{0.2cm}p{0.6cm}p{0.8cm}}
			\hline
			& 	\multicolumn{6}{c}{With color histogram} & \multicolumn{6}{|c}{Without color histogram} \\ \hline
			Method &	IDF1 &IDP &IDR	&	IDs	&	MOTA	&	MOTP &	IDF1 &IDP &IDR	&	IDs	&	MOTA	&	MOTP \\ \hline
			VGG &	72.9&67.1&79.8	&	14	&	75.8	&	82.1 & 65.9&60.8&	71.8&	24	&	75.4	&	\textbf{82.1}	\\
			ResNet &65.9&62.7&	69.3	&	35	&	83	&	80.9& 47.7&45.5	& 50.1	&	67	&	82.9	&	80.9	\\
			FCN7 & 35.6	&35.5&35.6 & 124	&	61.7	&	72.3 & 31.8	&31.8& 31.8 &	181	&	60.8	&	72.3	\\
			YOLO &67.1&61.8	&73.4	&	19	&	75.6	&	\textbf{81.6} & 69.5	& 63.9&	76&	28	&	75.6	&	81.5	\\ \hline
			Ours &\textbf{85.3}	&\textbf{86.2}&	\textbf{84.5}&	\textbf{10}	&	\textbf{93.4}	&	80.8 &\textbf{80.9}&\textbf{81.6}	& \textbf{80.1}	&	\textbf{17}	&	\textbf{93.2}	&	80.8	\\ \hline
		\end{tabular}
	}
	\end{center}
\end{table}

The performance of different approaches on Env 1 is shown in \tablename{~\ref{tab:tracking:env1}}.
Overall, our approach gives the best performance except the slightly worse MOTP in one test.
Env 2 has a much smaller set of training samples. And, the environment is more challenging due to the cluttered background (shelves, boxes, chairs, products).
The tracking results on Env 2 are shown in \tablename{~\ref{tab:tracking:env2}}. 
Our proposed method gives much better results than other baselines.
In particular, YOLO gives much poor performance due to the lack of training samples. FCN7 may be sufficient for crowd counting, but is poor for tracking.
\begin{table}[!tbp]
	\begin{center}
		\caption{Comparisons between different approaches for Env 2. }
		\label{tab:tracking:env2}
		\begin{tabular}{l|p{0.4cm}p{0.4cm}p{0.4cm}p{0.2cm}p{0.6cm}p{0.8cm}|p{0.4cm}p{0.4cm}p{0.4cm}p{0.2cm}p{0.6cm}p{0.8cm}}
			\hline
			& \multicolumn{6}{c}{With color histogram} & \multicolumn{5}{|c}{Without color histogram}  \\ \hline
			Method &	IDF1 & IDP & IDR	&	IDs	&	MOTA&MOTP &IDF1&IDP &IDR	&	IDs	&	MOTA	&	MOTP	\\ \hline
			VGG	&47.5&46.8	&48.2 &	37	&	77.6	&	80.7 	& 37.7&37.1	& 38.3	&	63	&	77.6	&	80.6	\\
			ResNet  & 55.7&54.8	&56.7&	36	&	79.4	&	82	& 30.3 & 29.3&	31.3 	&	64	&	74	&	82	\\
			FCN7 & 19.9	& 22.2	&18.1&	95	&	13.4	&	69.4& 15.1 & 16.8	& 13.8	&	124	&	11.7	&	69.2	\\
			YOLO & 10.5	& 8.9	&12.8&	78	&	-67.4	&	68 & 13.2&11.2&	16.1&	86	&	-69.7	&	68.3	\\ \hline
			Ours &  \textbf{90.4}&\textbf{90.9}&	\textbf{89.9}&	\textbf{3}	&	\textbf{97.1}	&	\textbf{86.5}&\textbf{88.4}& \textbf{88.9}	& \textbf{87.9}&	\textbf{5}	&	\textbf{97.1}	&	\textbf{86.5}	\\
			\hline
		\end{tabular}
	\end{center}
\end{table}

\subsection{Further discussions}
The experimental results demonstrate the effectiveness and the robustness of our 3D Deep Multi-Camera Tracking pipeline.
DMCT can be deployed into a real-time system.
We use a master computer connected to different cameras and
deploy the tracker using LibTorch (\url{https://pytorch.org/get-started/locally/}).
Our system can run in real time and one GeForce RTX 2080 Ti Graphics Card can drive eight connected cameras with 15 frames per second 
tracking frame rate.

\begin{figure}[!htbp]
	\begin{center}
			\includegraphics[width=0.975\linewidth]{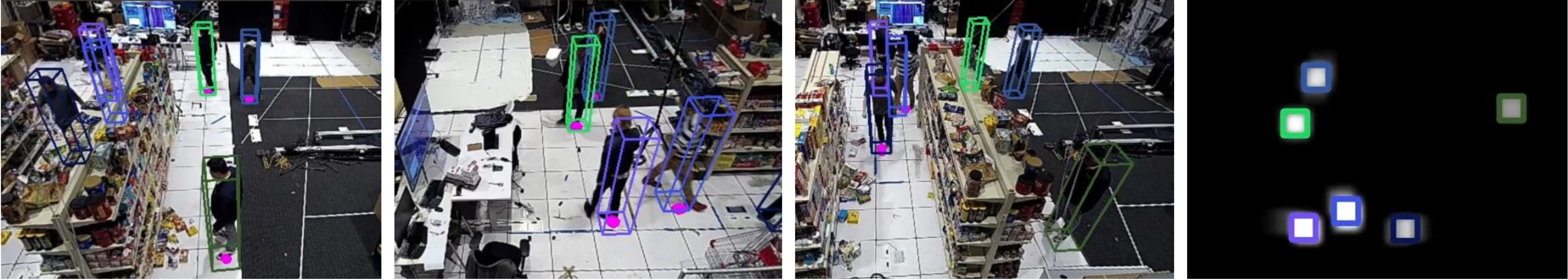}
	\end{center}
	\caption{Example images from a new environment with three network cameras (best viewed in color).}
	\label{fig:studiox:ips}
\end{figure}

Furthermore, we also construct another real-time system, where three network cameras are installed. 
We train our model using the \textbf{pseudo} ground truth from CrowdHuman (\url{https://www.crowdhuman.org}).
In particular, we use the center of the bottom edge of their ``full'' body bounding boxes as the ``ground point'' to train our model.
Surprisingly, the model generalizes very well and can be used to track people accurately using network cameras. \figurename{~\ref{fig:studiox:ips}} shows an example result of our real-time system.
The first three RGB images are the camera views with purple dots representing the detected ``ground points'' in each view. The 3D bounding boxes
are the projection of the tracking results (last image) on to each view with a fixed height.

\section{Conclusion}
In this work, we propose a real-time tracking pipeline (DMCT) using multiple RGB cameras, where each submodule is constructed
individually and the whole system is end-to-end in runtime.
The view-wise visual content is processed by a perspective-aware Deep GroundPoint Network, and is further fused together using 3D geometry.
Next, we propose a light-weight people detector on the fused heatmaps with Deep Glimpse Network as the person recognition model.
This design significantly simplifies our tracker. Our online tracker outperforms the state-of-the-art offline trackers on WILDTRACK dataset.
Furthermore, we also collect two multi-camera tracking datasets from two different environments.
The results on the two datasets further confirm the effectiveness and robustness of the proposed pipeline.
The flexibility of our method is also verified by a real-time system trained using the noisy ground point labels from CrowdHuman dataset.
Our proposed pipeline is ready to be deployed in real-world applications for robust and accurate multi-camera people tracking.

\clearpage
%
%
\bibliographystyle{plain}
\bibliography{egbib}

\begin{thebibliography}{10}

\bibitem{alahi2012freak}
Alexandre Alahi, Raphael Ortiz, and Pierre Vandergheynst.
\newblock Freak: Fast retina keypoint.
\newblock In {\em 2012 IEEE Conference on Computer Vision and Pattern
  Recognition}, pages 510--517. Ieee, 2012.

\bibitem{bai2018empirical}
Shaojie Bai, J~Zico Kolter, and Vladlen Koltun.
\newblock An empirical evaluation of generic convolutional and recurrent
  networks for sequence modeling.
\newblock {\em arXiv preprint arXiv:1803.01271}, 2018.

\bibitem{epfl2}
Pierre Baqu{\'e}, Fran{\c{c}}ois Fleuret, and Pascal Fua.
\newblock Deep occlusion reasoning for multi-camera multi-target detection.
\newblock In {\em Proceedings of the IEEE International Conference on Computer
  Vision}, pages 271--279, 2017.

\bibitem{singletrack1}
Quin Cai and Jake~K Aggarwal.
\newblock Automatic tracking of human motion in indoor scenes across multiple
  synchronized video streams.
\newblock In {\em Sixth International Conference on Computer Vision (IEEE Cat.
  No. 98CH36271)}, pages 356--362. IEEE, 1998.

\bibitem{chavdarova2018wildtrack}
Tatjana Chavdarova, Pierre Baqu{\'e}, St{\'e}phane Bouquet, Andrii Maksai, Cijo
  Jose, Timur Bagautdinov, Louis Lettry, Pascal Fua, Luc Van~Gool, and
  Fran{\c{c}}ois Fleuret.
\newblock Wildtrack: A multi-camera hd dataset for dense unscripted pedestrian
  detection.
\newblock In {\em Proceedings of the IEEE Conference on Computer Vision and
  Pattern Recognition}, pages 5030--5039, 2018.

\bibitem{duan2019centernet}
Kaiwen Duan, Song Bai, Lingxi Xie, Honggang Qi, Qingming Huang, and Qi~Tian.
\newblock Centernet: Keypoint triplets for object detection.
\newblock {\em arXiv preprint arXiv:1904.08189}, 2019.

\bibitem{epfl1}
Francois Fleuret, Jerome Berclaz, Richard Lengagne, and Pascal Fua.
\newblock Multicamera people tracking with a probabilistic occupancy map.
\newblock {\em IEEE transactions on pattern analysis and machine intelligence},
  30(2):267--282, 2007.

\bibitem{centroid}
Dirk Focken and Rainer Stiefelhagen.
\newblock Towards vision-based 3-d people tracking in a smart room.
\newblock In {\em Proceedings. Fourth IEEE International Conference on
  Multimodal Interfaces}, pages 400--405. IEEE, 2002.

\bibitem{Girshick_2014_CVPR}
Ross Girshick, Jeff Donahue, Trevor Darrell, and Jitendra Malik.
\newblock Rich feature hierarchies for accurate object detection and semantic
  segmentation.
\newblock In {\em The IEEE Conference on Computer Vision and Pattern
  Recognition (CVPR)}, June 2014.

\bibitem{he2016deep}
Kaiming He, Xiangyu Zhang, Shaoqing Ren, and Jian Sun.
\newblock Deep residual learning for image recognition.
\newblock In {\em Proceedings of the IEEE conference on computer vision and
  pattern recognition}, pages 770--778, 2016.

\bibitem{medioni}
Jinman Kang, Isaac Cohen, Gerard Medioni, et~al.
\newblock Tracking people in crowded scenes across multiple cameras.
\newblock In {\em Asian conference on computer vision}, volume~7, page~15.
  Citeseer, 2004.

\bibitem{ucf}
Saad~M Khan and Mubarak Shah.
\newblock A multiview approach to tracking people in crowded scenes using a
  planar homography constraint.
\newblock In {\em European Conference on Computer Vision}, pages 133--146.
  Springer, 2006.

\bibitem{law2018cornernet}
Hei Law and Jia Deng.
\newblock Cornernet: Detecting objects as paired keypoints.
\newblock In {\em Proceedings of the European Conference on Computer Vision
  (ECCV)}, pages 734--750, 2018.

\bibitem{opt1}
Laura Leal-Taixe, Gerard Pons-Moll, and Bodo Rosenhahn.
\newblock Branch-and-price global optimization for multi-view multi-target
  tracking.
\newblock In {\em 2012 IEEE Conference on Computer Vision and Pattern
  Recognition}, pages 1987--1994. IEEE, 2012.

\bibitem{leutenegger2011brisk}
Stefan Leutenegger, Margarita Chli, and Roland Siegwart.
\newblock Brisk: Binary robust invariant scalable keypoints.
\newblock In {\em 2011 IEEE international conference on computer vision
  (ICCV)}, pages 2548--2555. Ieee, 2011.

\bibitem{larry}
A~Mittal and LS~Davis.
\newblock M2tracker: A multi-view approach to segmenting and tracking people in
  a cluttered scene.
\newblock 51(3), 2003.

\bibitem{mnih2014recurrent}
Volodymyr Mnih, Nicolas Heess, Alex Graves, et~al.
\newblock Recurrent models of visual attention.
\newblock In {\em Advances in neural information processing systems}, pages
  2204--2212, 2014.

\bibitem{occ}
Kazuhiro Otsuka and Naoki Mukawa.
\newblock Multiview occlusion analysis for tracking densely populated objects
  based on 2-d visual angles.
\newblock In {\em Proceedings of the 2004 IEEE Computer Society Conference on
  Computer Vision and Pattern Recognition, 2004. CVPR 2004.}, volume~1, pages
  I--I. IEEE, 2004.

\bibitem{papadimitrou1982combinatorial}
Christos~H Papadimitrou and Kenneth Steiglitz.
\newblock Combinatorial optimization: algorithms and complexity.
\newblock 1982.

\bibitem{yolo}
Joseph Redmon, Santosh Divvala, Ross Girshick, and Ali Farhadi.
\newblock You only look once: Unified, real-time object detection.
\newblock In {\em Proceedings of the IEEE conference on computer vision and
  pattern recognition}, pages 779--788, 2016.

\bibitem{ren2015faster}
Shaoqing Ren, Kaiming He, Ross Girshick, and Jian Sun.
\newblock Faster r-cnn: Towards real-time object detection with region proposal
  networks.
\newblock In {\em Advances in neural information processing systems}, pages
  91--99, 2015.

\bibitem{shao2018crowdhuman}
Shuai Shao, Zijian Zhao, Boxun Li, Tete Xiao, Gang Yu, Xiangyu Zhang, and Jian
  Sun.
\newblock Crowdhuman: A benchmark for detecting human in a crowd.
\newblock {\em arXiv preprint arXiv:1805.00123}, 2018.

\bibitem{simonyan2014very}
Karen Simonyan and Andrew Zisserman.
\newblock Very deep convolutional networks for large-scale image recognition.
\newblock {\em arXiv preprint arXiv:1409.1556}, 2014.

\bibitem{opt2}
Longyin Wen, Zhen Lei, Ming-Ching Chang, Honggang Qi, and Siwei Lyu.
\newblock Multi-camera multi-target tracking with space-time-view hyper-graph.
\newblock {\em International Journal of Computer Vision}, 122(2):313--333,
  2017.

\bibitem{margrit}
Zheng Wu, Nickolay~I Hristov, Tyson~L Hedrick, Thomas~H Kunz, and Margrit
  Betke.
\newblock Tracking a large number of objects from multiple views.
\newblock In {\em 2009 IEEE 12th International Conference on Computer Vision},
  pages 1546--1553. IEEE, 2009.

\bibitem{HED}
Saining Xie and Zhuowen Tu.
\newblock Holistically-nested edge detection.
\newblock In {\em Proceedings of the IEEE international conference on computer
  vision}, pages 1395--1403, 2015.

\bibitem{opt3}
Yuanlu Xu, Xiaobai Liu, Yang Liu, and Song-Chun Zhu.
\newblock Multi-view people tracking via hierarchical trajectory composition.
\newblock In {\em Proceedings of the IEEE Conference on Computer Vision and
  Pattern Recognition}, pages 4256--4265, 2016.

\bibitem{count}
Danny~B Yang, Leonidas~J Guibas, et~al.
\newblock Counting people in crowds with a real-time network of simple image
  sensors.
\newblock In {\em null}, page 122. IEEE, 2003.

\bibitem{yu2018deep}
Fisher Yu, Dequan Wang, Evan Shelhamer, and Trevor Darrell.
\newblock Deep layer aggregation.
\newblock In {\em Proceedings of the IEEE Conference on Computer Vision and
  Pattern Recognition}, pages 2403--2412, 2018.

\bibitem{count19}
Qi~Zhang and Antoni~B Chan.
\newblock Wide-area crowd counting via ground-plane density maps and multi-view
  fusion cnns.
\newblock In {\em Proceedings of the IEEE Conference on Computer Vision and
  Pattern Recognition}, pages 8297--8306, 2019.

\bibitem{zhou2019objects}
Xingyi Zhou, Dequan Wang, and Philipp Kr{\"a}henb{\"u}hl.
\newblock Objects as points.
\newblock In {\em arXiv preprint arXiv:1904.07850}, 2019.

\end{thebibliography}

\end{document}